\documentclass[sigconf]{acmart}

\AtBeginDocument{%
  \providecommand\BibTeX{{%
    \normalfont B\kern-0.5em{\scshape i\kern-0.25em b}\kern-0.8em\TeX}}}

\setcopyright{acmcopyright}

\copyrightyear{2020}
\acmYear{2020}
\setcopyright{acmlicensed}\acmConference[MM '20]{Proceedings of the 28th ACM International Conference on Multimedia}{October 12--16, 2020}{Seattle, WA, USA}
\acmBooktitle{Proceedings of the 28th ACM International Conference on Multimedia (MM '20), October 12--16, 2020, Seattle, WA, USA}
\acmPrice{15.00}
\acmDOI{10.1145/3394171.3413815}
\acmISBN{978-1-4503-7988-5/20/10}

\acmSubmissionID{688}

\usepackage{docmute}
\graphicspath{figs/}

\begin{document}

\title{Surpassing Real-World Source Training Data: Random 3D Characters for Generalizable Person Re-Identification}


\author{Yanan Wang}
\email{yanan.wang@inceptioniai.org}
\author{Shengcai Liao}\authornote{Shengcai Liao is the Corresponding Author.}
\email{scliao@ieee.org}
\affiliation{%
 \institution{Inception Institute of Artificial Intelligence (IIAI)}
 \city{Masdar City}
 \state{Abu Dhabi}
 \country{UAE}
}
\author{Ling Shao}
\affiliation{%
 \institution{Inception Institute of Artificial Intelligence (IIAI)}
 \city{Masdar City}
 \state{Abu Dhabi}
 \country{UAE}
}
\affiliation{%
 \institution{Mohamed bin Zayed University of Artificial Intelligence}
 \city{Masdar City}
 \state{Abu Dhabi}
 \country{UAE}
}
\email{ling.shao@ieee.org}


\begin{abstract}
Person re-identification has seen significant advancement in recent years. However, the ability of learned models to generalize to unknown target domains still remains limited. One possible reason for this is the lack of large-scale and diverse source training data, since manually labeling such a dataset is very expensive and privacy sensitive. To address this, we propose to automatically synthesize a large-scale person re-identification dataset following a set-up similar to real surveillance but with virtual environments, and then use the synthesized person images to train a generalizable person re-identification model. Specifically, we design a method to generate a large number of random UV texture maps and use them to create different 3D clothing models. Then, an automatic code is developed to randomly generate various different 3D characters with diverse clothes, races and attributes. Next, we simulate a number of different virtual environments using Unity3D, with customized camera networks similar to real surveillance systems, and import multiple 3D characters at the same time, with various movements and interactions along different paths through the camera networks. As a result, we obtain a virtual dataset, called RandPerson, with 1,801,816 person images of 8,000 identities. By training person re-identification models on these synthesized person images, we demonstrate, for the first time, that models trained on virtual data can generalize well to unseen target images, surpassing the models trained on various real-world datasets, including CUHK03, Market-1501, DukeMTMC-reID, and almost MSMT17. The RandPerson dataset is available at \url{https://github.com/VideoObjectSearch/RandPerson}.

\end{abstract}



\begin{teaserfigure}
  \centering
  \includegraphics[width=1\textwidth]{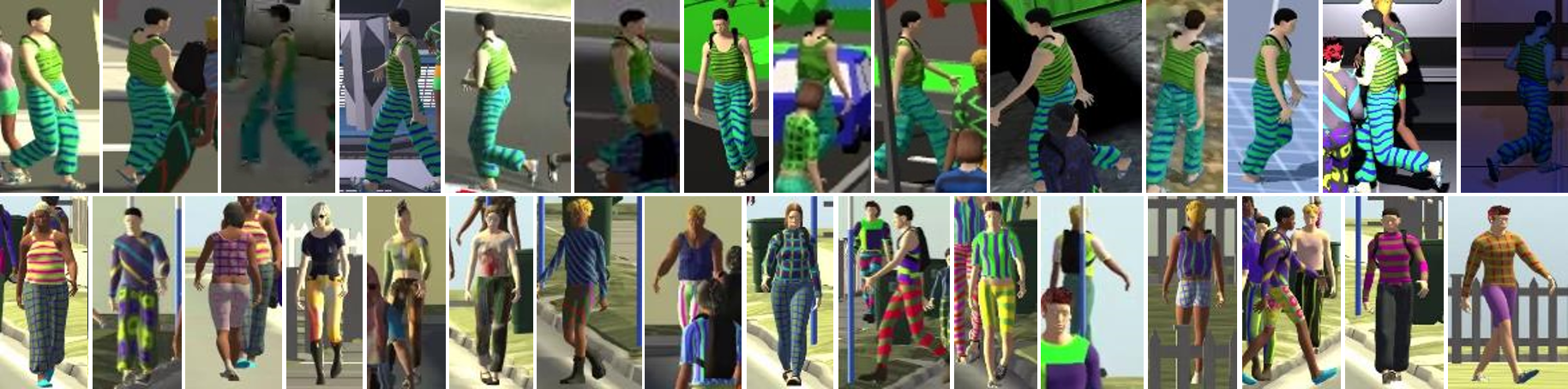}
  \caption{Sample images from the proposed RandPerson dataset, with the same character in different scenes (first row), and different characters in the same scene (second row). This is the first synthetic person re-identification dataset with a set-up similar to real video surveillance systems, i.e. with a camera network and multiple people moving at the same time. The dataset contains 1,801,816 synthesized person images of 8,000 identities. Images in this dataset generally contain different viewpoints, poses, illuminations, backgrounds, occlusions, and resolutions.}
  \Description{}
  \label{fig:sample_from_RandPerson}
\end{teaserfigure}

\maketitle

\section{Introduction}
Person re-identification, which aims at retrieving a query image person from a large number of gallery images, has been significantly advanced in recent years \cite{ye2020deep}. It is a hot research topic in both computer vision \cite{ye2020deep} and multimedia retrieval \cite{fan2018unsupervised, liu2016multi, wang2018learning}. The most effective state-of-the-art algorithms employ supervised learning and have achieved excellent results in within-dataset evaluation \cite{gong2014re,zheng2016person,ye2020deep}. However, the generalization of these learned models to unknown target domains remains poor \cite{hu2014cross,yi2014deep,Liao-ECCV2020-QAConv}. This is largely a result of the present lack in large-scale and diverse source training datasets, since acquiring and manually labeling such datasets is highly expensive and privacy sensitive. The current largest person re-identification dataset MSMT17 \cite{wei2018person} contains 126,441 images of 4,101 identities, but they were all collected on the same campus. Thus, it still remains challenging to find real-world data that is diverse in various aspects, such as scenes, illumination, clothes, etc.

To address this, several recent works have started to use game engines to render diverse person images. For example, SOMAset \cite{barbosa2018looking} contains 50 3D characters each with 11 types of outfits, targeting at cloth-independent person re-identification. SyRI \cite{bak2018domain} includes 100 3D characters created under multiple HDR environment maps, which simulate realistic indoor and outdoor lighting. PersonX \cite{sun2019dissecting} contains 1,266 3D characters shown with various viewpoints, poses, illuminations, and backgrounds. However, all these works used publicly available 3D characters or hand-crafted models, limiting their scalability in terms of the number of synthesized people. Besides, data was captured for one character at a time using a single camera, which is far from real surveillance settings.

In this paper, we propose to automatically synthesize a large-scale person re-identification dataset following a set-up similar to real surveillance but with virtual environments, and then use the synthesized person images to train a generalizable person re-identification model. Different from previous works using public 3D characters or hand-crafted models, we design a method to generate a large number of random UV texture maps \cite{chang2006modeling} and use them to create different 3D clothing models. Then, an automatic code is developed to randomly generate various different 3D characters with diverse clothes, races and attributes. This makes the number of identities much more scalable. Then, we simulate a number of diverse virtual environments using Unity3D \cite{unity}, with customized camera networks similar to real surveillance systems, and import multiple 3D characters at the same time, with various movements and interactions along different paths through the camera networks. The virtual environments include both indoor and outdoor scenes and range from bright conditions full of sunshine to gloomy settings, and from cities to country fields.

As a result, by the above multimedia computer graphics method we obtain a virtual dataset, called RandPerson, with 1,801,816 person images of 8,000 identities. Fig. \ref{fig:sample_from_RandPerson} shows some sample images of RandPerson. By training person re-identification models on these synthesized person images, we demonstrate, for the first time, that models trained on virtual data can generalize well to unseen target images, surpassing those trained on various real-world datasets, including CUHK03 \cite{li2014deepreid}, Market-1501 \cite{zheng2015scalable}, DukeMTMC-reID \cite{zheng2017unlabeled}, and almost MSMT17 \cite{wei2018person}. Besides, combining the RandPerson and real-world datasets as multimedia data for training further improves the results, leading to the best performance in cross-database person re-identification compared to other configurations.

\section{Related Work}
Person re-identification has seen significant progress in recent years as a result of deep learning \cite{ye2020deep}. However, due to the limited amount of labeled data and the diversity of real-world surveillance, the generalization ability of person re-identification models is poor in unseen scenarios. To address this, many transfer learning or domain adaptation approaches have been proposed to obtain better results \cite{ye2020deep}. However, these come with the heavy cost of requiring additional training in the target domain.

Many person re-identification datasets have been created, including both real-world and synthetic. For real-world images, the widely used datasets include VIPeR \cite{gray2008viewpoint}, iLIDS \cite{zheng2009associating}, GRID \cite{loy2013person}, PRID2011 \cite{hirzer2011person}, CUHK01-03 \cite{li2012human,li2013locally,li2014deepreid}, Market-1501 \cite{zheng2015scalable}, DukeMTMC-
reID \cite{zheng2017unlabeled}, Airport \cite{gou2018systematic}, and MSMT17 \cite{wei2018person}. The Airport dataset has 39,902 images of 9,651 identities, which makes it large in terms of identities but with a limited number of images per subject. The current largest dataset MSMT17 contains 126,441 images of 4,101 identities captured from 15 cameras. However, most of these datasets were collected on a single campus. It thus still remains challenging to find real data that is diverse in various aspects, such as scene, background, illumination, viewpoint, clothes, etc.

As for synthetic datasets, Barbosa \textit{et al.} proposed the SOMAset \cite{barbosa2018looking}, which contains 50 3D characterss, each with 11 types of outfits, targeting at cloth-independent person re-identification. It has 100,000 bounding boxes taken with 250 different camera orientations in a single scene. Bak \textit{et al.} proposed the SyRI dataset \cite{bak2018domain}, which contains 3D characters rendered under different illumination conditions to enrich the diversity. It has 1,680,000 bounding boxes of 100 identities. Sun \textit{et al.} proposed the PersonX dataset \cite{sun2019dissecting}, which contains 1,266 3D characters rendered in Unity3D, and 273,456 bounding boxes taken with six cameras. However, all of these works used publicly available 3D characters or hand-crafted models, limiting their scalability in terms of the possible number of synthesized persons. Furthermore, their images only provide one character at a time, under a single camera setting. In contrast, this work provides greater scalability by automatically generating a large number of 3D characters diverse in dress and attributes, and renders them simultaneously using camera networks similar to real surveillance systems. We use Unity3D to render characters following PersonX, but explore much more in running multiple persons and multiple cameras simultaneously.

\section{Random 3D Characters}
\subsection{3D Models in MakeHuman}

MakeHuman \cite{makehuman} is a free and open-source software used to create 3D character models. It also has its own extensive community where designers can contribute their models, including a wide variety of clothes, hairstyles, beards, etc., which can be used to extend the resources for developing new characters.

\begin{figure}[htbp]
\centering
\includegraphics[scale=0.22]{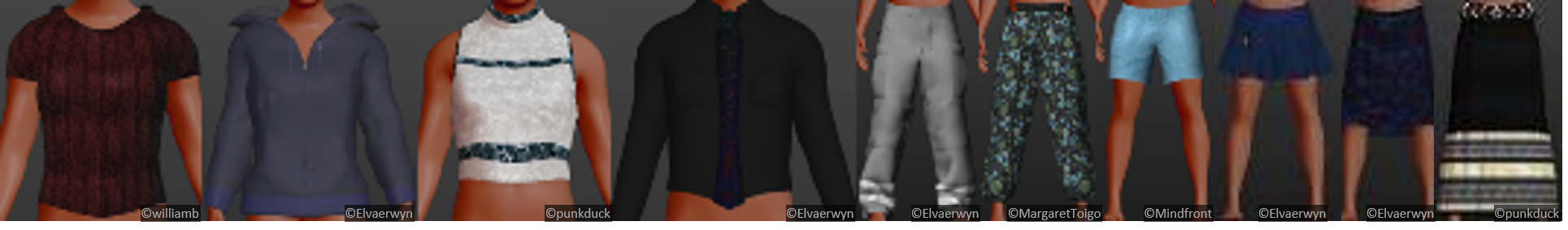}
\caption{Different types of clothing in MakeHuman, including shirts, pants and skirts.}
\label{fig:clothing_types}
\end{figure}

\begin{figure}[ht]
\centering
\includegraphics[scale=0.2]{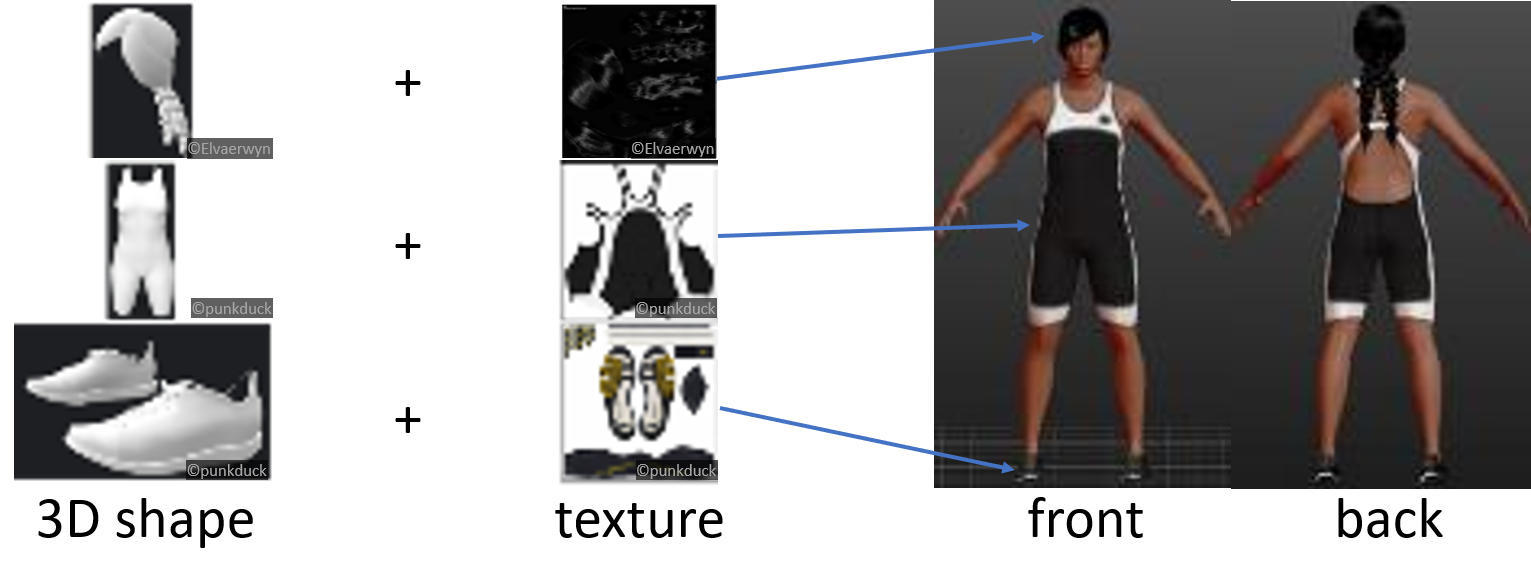}
\caption{Illustration of how to create a 3D character in MakeHuman, with 3D shape models of clothing and accessories (left), and their corresponding UV texture maps (middle). A character is created by applying these models onto a standard 3D human model in MakeHuman (right).}
\label{fig:character_create}
\end{figure}

Characters in MakeHuman are built upon a standard 3D body shape, using various body extensions (e.g. hairstyles and beards), clothes, and accessories. Clothes models include most ordinary types of clothing, such as dresses, suits, shirts, coats, jeans, trousers, shorts, and skirts. Fig. \ref{fig:clothing_types} shows some examples of different types of clothing. Each clothing model contains a 3D shape model, and a corresponding UV texture map of the model, as shown in the first two columns of Fig. \ref{fig:character_create}. The UV texture map is used to paint the 3D shape model via the UV mapping operation.

In MakeHuman, new 3D characters are created through simple drag and drop operations in the working panel, so as to combine various components together and adjust their attributes, such as skeleton, body parameters, clothes, etc. Fig. \ref{fig:character_create} shows an example of how to create a 3D character in MakeHuman. After UV mapping, the textured 3D components can be applied to the standard 3D human model to form a new individualized 3D character.

\subsection{Generating Random Clothing}

The MakeHuman community provides some clothing models, but there are only a few hundred with useful ordinary clothes. For our project, we download 179 clothing models from the community. This limits the number of characters that can be created. However, since each clothing model comes with a UV texture map defining the clothing texture, altering this can significantly change the appearance of the clothing. Therefore, in this paper, we propose to generate a large number of clothing models by altering the UV texture maps of available clothing models.

\begin{figure}[ht]
\centering
\includegraphics[scale=0.22]{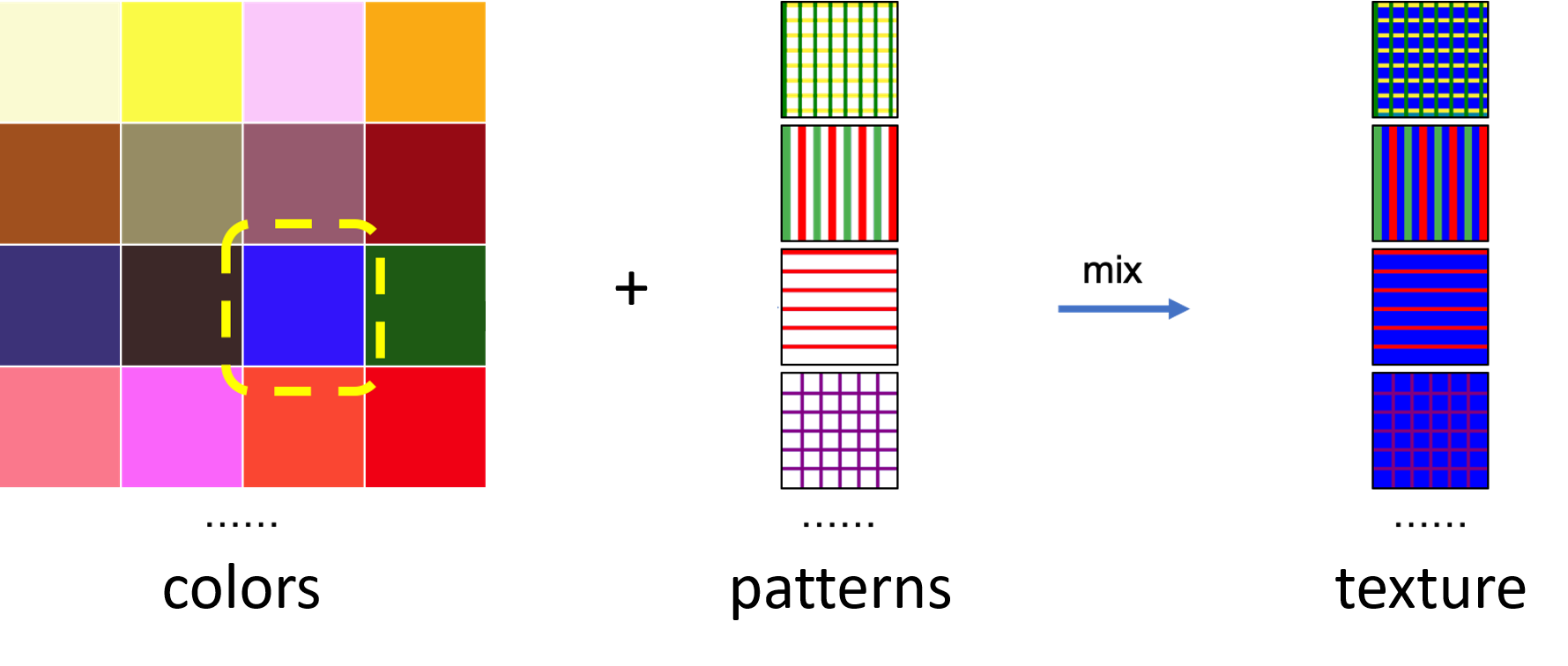}
\caption{Illustration of how to generate UV texture maps. A color palette (left) is created, with 625 colors uniformly sampled from the HSV space. Besides, 16 additional texture patterns (middle) are generated, which include different stripes and spots. Combining different colors and texture patterns results in $625\times16=10,000$ UV texture maps (right).}
\label{fig:UV_texture_maps_generate}
\end{figure}

We generate new UV texture maps in two ways. The first is to search and download images from the Internet, and use them directly as new UV texture maps. Accordingly, 5,772 images are downloaded, including landscapes, animals, and texture patterns. However, most of these images are complicated, and the color distribution cannot be controlled. Therefore, alternatively, we design a method to generate random UV texture maps. Firstly, as shown in Fig. \ref{fig:UV_texture_maps_generate}, a color palette is generated, which contains representative colors sampled from the HSV space. This is done by using a regular grid in the HSV space, with a step of 15 in H, 0.2 in S and V, resulting in $24\times5\times5=600$ colors. However, these do not contain gray samples. Therefore, additionally, gray samples with H = 0, S = 0, and 25 values evenly distributed on V are sampled, so together there are 625 representative colors composing the color palette. 
In addition, 16 texture patterns are generated, which include different spots, stripes and lattices. By sampling a background color from the color palette, and then drawing a texture pattern on the background image, a new UV texture map is generated, which can make clothes look spotted, striped or plaid. Combining different colors and texture patterns results in $625\times16=10,000$ UV texture maps.

\begin{figure}[ht]
\centering
\includegraphics[scale=0.3]{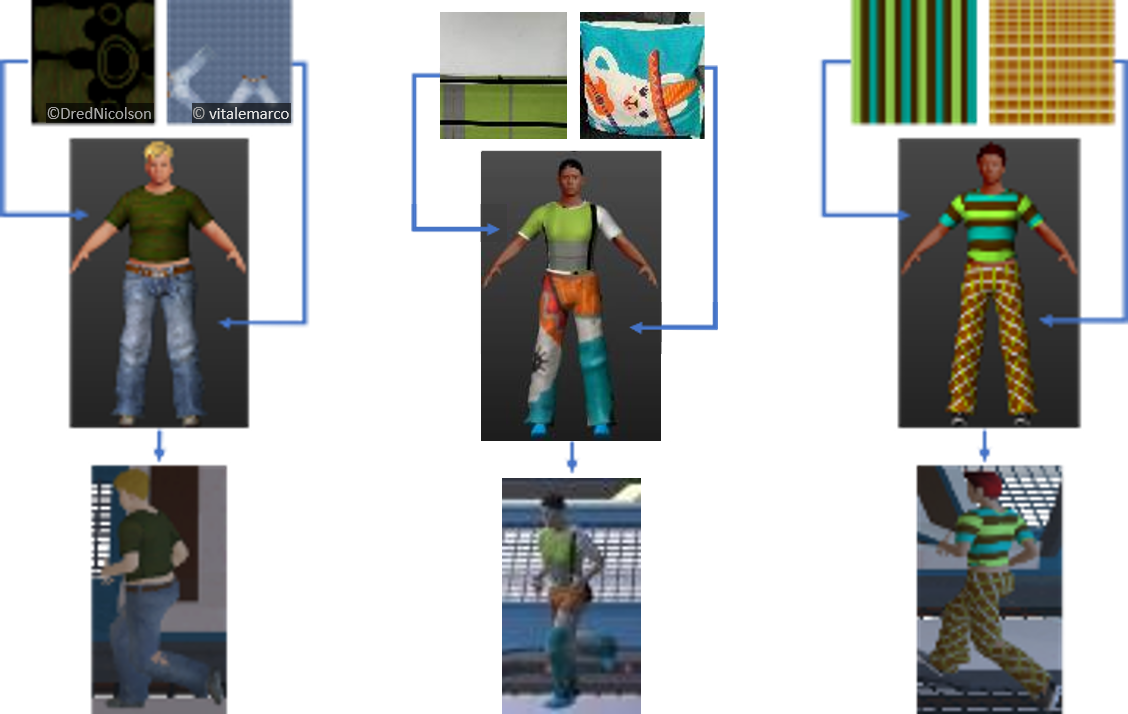}
\caption{Examples of generated clothes. From top to bottom: UV texture maps, 3D characters wearing the new clothing after UV mapping, images rendered in a Unity scene. From left to right: 3D characters generated using the clothing models with the original UV texture map, a UV map with a web image, and a UV map with a random color and texture pattern.}
\label{fig:generated_clothes_illustration}
\end{figure}

Now that we have a great number of UV texture maps, a new clothing model can be created by choosing an existing model, and replacing its UV texture map with one random sample from the pool of our downloaded or generated UV texture maps. Examples of generated clothing can be seen in Fig. \ref{fig:generated_clothes_illustration}, which shows how different UV texture maps can be used to texturize the same clothing model, and how the generated clothes and rendered images look differently. By utilizing these large numbers of UV texture maps, we obtain a series of different outfits, which can be further used to create different 3D characters.

\subsection{Creating Random 3D Characters}

In MakeHuman, there is a standard human model that can be built upon by varying several attributes, such as skeleton, body parameters, and clothes. The body parameters, such as height, weight, waistline, etc., can be used to change the overall 3D body look. In addition, features such as the face, hands, neck, gender, age, and skin color can also be adjusted. Finally, various clothes, as well as accessories such as shoes, hairstyles, beards, and hats, can also be used to develop different characters.

Note that MakeHuman only supports interactive operations manually in wearing different components together. However, it is inefficient to create each character one by one, by hand, and remembering which components have already been used is not an easy task. Therefore, after analyzing the exported file structure (.mhm file) of the 3D characters created in MakeHuman, we developed a Python code to automatically generate a large number of new characters. Specifically, all clothing models are categorized into several types, such as female clothes, male clothes, and universal. A character is created starting from the standard human model, which is further adjusted step by step, including combining different components and setting different attributes or parameters. In the generation process, we set an equal probability of producing men and women. Different skin colors are equally distributed among the available types. The ages are uniformly distributed in [20,90]. The weights are uniformly distributed in the range allowed by the system. The heights of the character follow a Gaussian distribution $\mathcal{N}(\mu, \sigma)$, where the mean $\mu=170 cm$ and the standard deviation $\sigma=5.7 cm$ for men, while $\mu=160 cm$ and $\sigma=5.2 cm$ for women. Then, we randomly sample a clothing model for each new character according to its gender. Furthermore, we randomly add accessories to different characters, such as beards for men and necklaces and bows for women. Shoes and hairstyles are randomly appended according to the gender of the character. Hats and backpacks are also randomly equipped to both male and female characters. 

Finally, we created 8,000 different 3D characters, including 114 characters with the original clothing models, 2,886 with UV texture maps from web images, and 5,000 with random UV maps. The character models created are saved as standard MakeHuman .mhm files, and can then be exported to several other file types to be used in animation software. In this work, we use the .fbx file type for the exported 3D character model.

\section{Diverse Simulations in Unity3D}
\subsection{Programming in Unity3D}

Unity3D \cite{unity} is a cross-platform 3D game engine. It has a well-established asset store where various useful resources are available, such as scenes, characters, actions, and objects. However, the free 3D characters from the asset store are limited to only a few hundreds, therefore, the random 3D characters we designed are used instead.

Unity3D provides various different environments and camera positions, allowing observation of different scenarios. Besides, it provides a recording feature, so that we can create a video of the imported characters' movements in the virtual environment, and track their positions.

Scripts can be added to Unity's objects so that they move according to the program settings. For example, the environmental lighting can be changed as needed to simulate dynamic variations. Besides, once 3D characters are imported to Unity3D, they can be animated to walk, run, sit, etc. Furthermore, we can also use scripts to control the movement of characters from one point to another to create pedestrian data where the characters walk randomly.

\subsection{Customization of Virtual 3D Scenes}

The Unity3D Asset Store provides a variety of scenes, including outdoor environments such as deserts, forests, streets, cities and villages, and indoor environments such as gyms, bedrooms and living rooms. We can also use a 3D model to build a new scene and add different lighting forms, such as parallel light, point light, and various colors of light, to change the background appearance and environmental illumination.

We downloaded several environments from the Unity3D Asset Store, including eight outdoor and three indoor scenes. To simulate different factors affecting person re-identification, including variations in viewpoint, lighting, background, and distance, the downloaded scenes are modified as follows by adding new scripts. First, different walkable places are set for each scene. Second, the position, angle, and intensity of the lighting are set to change regularly. We vary the light intensity from 0 to 1.5, and gradually change the position and direction of the light, so as to simulate the change in lighting that occurs throughout the day. Furthermore, some colored light is also added to change the chrominance of the scene. 

\begin{figure}[ht]
\centering
\includegraphics[scale=0.44]{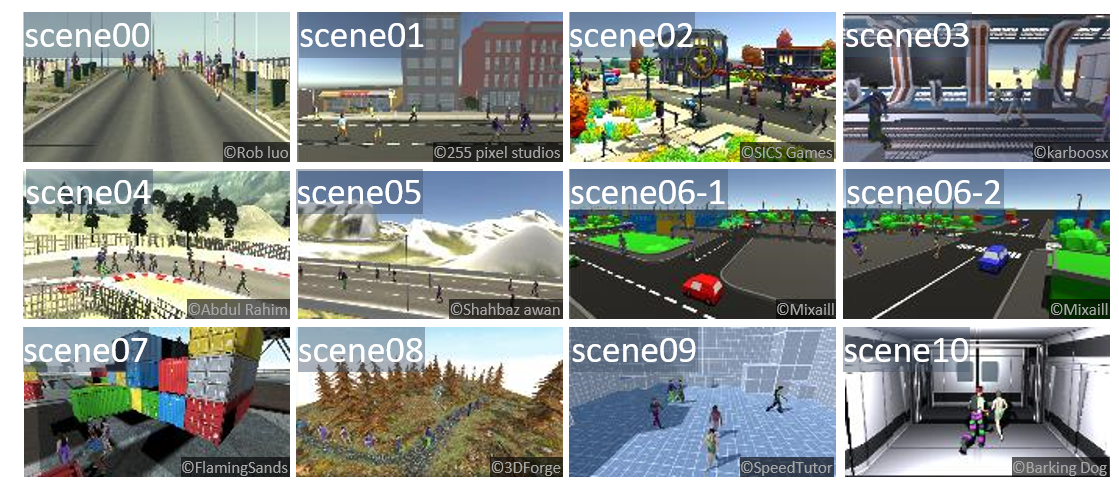}
\caption{Examples of custom Unity3D scenarios produced in this work.}
\label{fig:scenarios_example}
\end{figure}

Finally, we obtain a set of customized environments, comprising streets, forest paths, highways and laboratories, among others, with bright light, dark light, blue light, etc. Fig. \ref{fig:scenarios_example} shows some example scenarios used in this work.

\begin{figure*}[ht]
\centering
\includegraphics[width=0.65\textwidth]{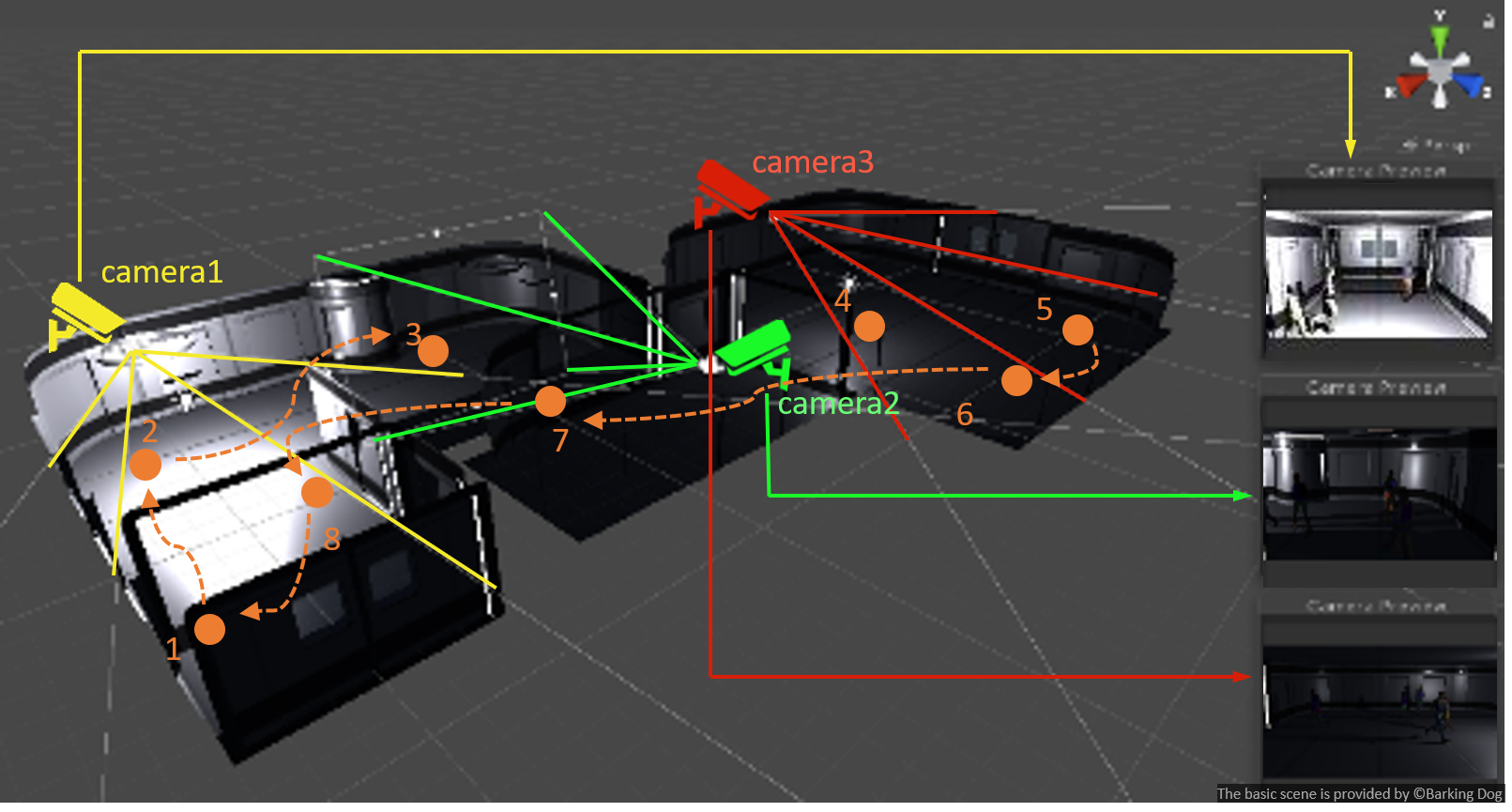}
\caption{Example configuration of camera networks and character movements. Three cameras are mounted, shown in different colors. Straight lines represent each camera's field of view, while arrows point to the corresponding examples of snapshots. Orange dots represent the destination points we set, and the dotted lines represent a possible movement route.}
\label{fig:camera_configuration}
\end{figure*}

\subsection{Configuration of Camera Networks and Character Movements}

Different from previous works using only one camera in simulation, in this work, we set up a network with multiple cameras for each scene, and run them simultaneously in the virtual environments, simulating real video surveillance. Since viewpoint, background and resolution are essential factors for person re-identification, we choose areas with backgrounds that are as different as possible when setting up different cameras. Besides, the cameras use different heights, angles of view, distances, and visual ranges, so that the same person can be displayed in numerous different ways. An example is shown in Fig. \ref{fig:camera_configuration}, where three cameras are mounted, represented in yellow, green, and red, respectively, covering diverse fields of view. Furthermore, setting different camera distances generates person images in different resolutions. Considering the above principles, we set up 19 cameras in 11 scenes in total. Due to the differences in camera height, angle, and distance, the person images generated vary extensively in background, viewpoint, and resolution, enhancing the diversity of data (see Fig. \ref{fig:sample_from_RandPerson}).

Another difference from previous works is that the characters in this work move simultaneously through the camera network, mimicking real video surveillance. As a result, person-to-person occlusions commonly occur in the simulation, which is also an important factor in person re-identification. Specifically, to establish a character's movement route, we set a cyclic sequence of destination points, as shown in Fig. \ref{fig:camera_configuration}. Then, each imported character is randomly assigned a starting point in this sequence, and will move to the following destination points one by one. We use the path-finding system provided by Unity3D to navigate characters between two destination points. In the simulation, each character is imported into the scene in turn, with several seconds delay between characters, and they are then removed after visiting a certain number of destinations. The character importing is controlled by the maximum capacity of a scene. With the above settings, many characters will move along different paths in the scene, avoiding crowd gathering. Besides, the angles and positions of the captured characters are different, which further increases the variations in viewpoint and background.

As for poses, in addition to the variations due to different camera viewpoints, several animation methods are also applied, including three walking modes and three running modes. Accordingly, the cropped images of characters may appear in many stages of their movements, which significantly enriches the variety of poses.

\subsection{Video Capturing and Image Cropping}

To obtain more comprehensive data in addition to pedestrian images, the Unity Recorder component is applied to record videos of pedestrian movements. Videos from all cameras are recorded simultaneously as characters move together in a scene. The video recording is set to high-quality and 24 FPS, with a resolution of $1920\times1080$. To track each character's position and pose so that they can be used for automatic image cropping, along with a video an annotation file is also created to record keypoint locations of all characters during the simulation. This is achieved thanks to the skeleton data provided by the MakeHuman. The standard skeleton used in this work has 31 keypoints. In the simulation, to reduce file size, only seven keypoints are saved, including head, shoulders, hands, and feet.

\begin{figure}[ht]
\centering
\includegraphics[scale=0.25]{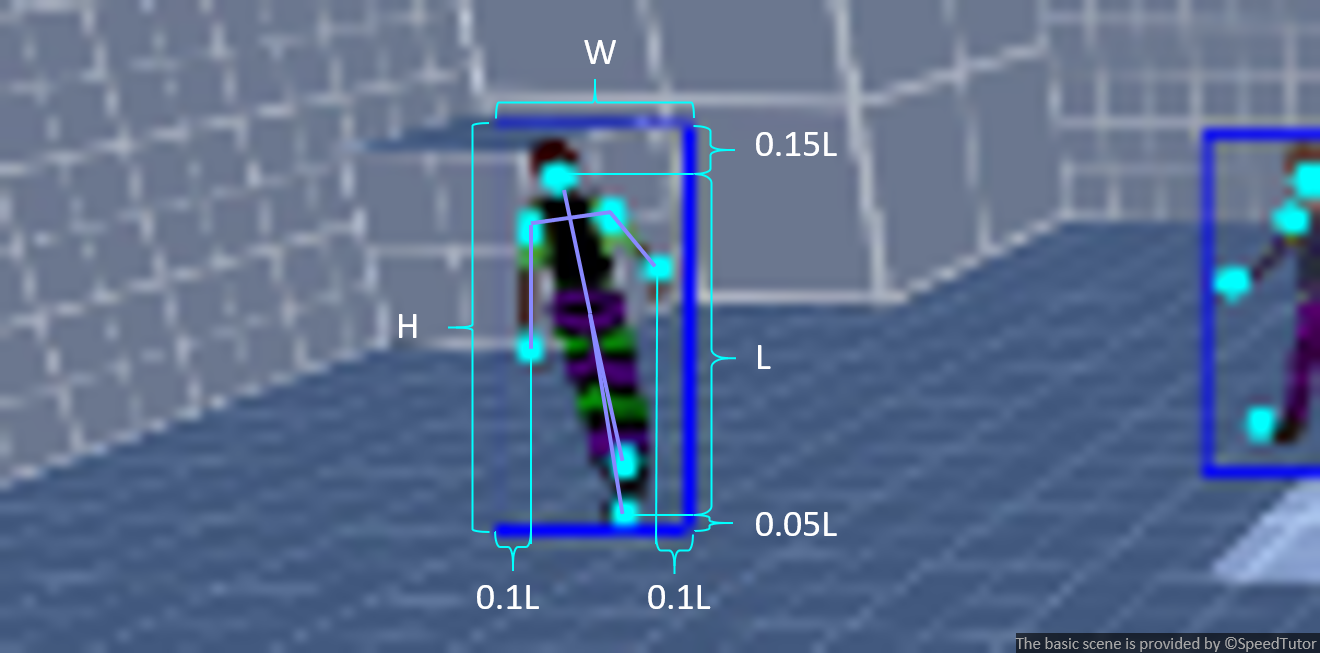}
\caption{Example of bounding box cropping.}
\label{fig:crop_images}
\end{figure}

To crop character images from videos, frames are first uniformly sampled from the recorded videos according to the speed of pedestrian movement in each video. For most videos, the sampling rate is every 0.5 seconds. Then, as illustrated in Fig. \ref{fig:crop_images}, for each character, the height $H$ is determined by the maximal vertical range (denoted as $L$) of keypoints, plus $0.15L$ over the uppermost keypoint and $0.05L$ under the lowermost keypoint. As for the width $W$, upon the maximal horizontal range of keypoints, $0.1L$ is appended on both the left and right sides. Finally, if $W/H<0.4$, an equal margin on both sides will further be appended to make $W/H=0.4$. Examples of cropped images can be seen in Fig. \ref{fig:sample_from_RandPerson}.


\begin{table*}[htbp] 
 \caption{Statistics of different synthetic datasets\protect\footnotemark. "Camera network" and "person interaction" indicate video recording with multiple cameras and multiple moving characters, respectively, simultaneously in the scene. The bounding boxes (bboxes) column shows the actual data used in the experiment, followed by the total amount of data.}
 \label{tab:dataset_compare}
 \begin{center}
  \begin{tabular}{|l|c|c|c|c|c|c|c|c|c|}
   \hline
   Dataset & \#Identities & \#Scenes & \#Cameras  & \#Videos & \#BBoxes & Camera Network & Person Interaction \\
   \hline\hline
   SOMAset \cite{barbosa2018looking}   & 50 & 1 & 250 & none &100,000 / 100,000& no & no\\
    SyRI \cite{bak2018domain}    & 100& 140&280&28,000&56,000 / 1,680,000&no& no\\
   PersonX \cite{sun2019dissecting}   & 1,266 &1& 6& none &273,456 / 273,456& no& no\\
   RandPerson& 8,000& 11& 19  & 38 &132,145 / 1,801,816& yes  & yes\\
   \hline
  \end{tabular}
 \end{center}
\end{table*}

\subsection{Summary of the RandPerson Dataset}

With the above procedures, we obtained a synthetic person re-identification dataset called RandPerson. It comprises synthetic person images with 8,000 identities, 38 videos, and 1,801,816 images. 
Fig. \ref{fig:sample_from_RandPerson} provides some examples from RandPerson. The first row shows the same character in different scenes, and the second row shows different characters in the same scene. As can be observed, the cropped images generally have different viewpoints, poses, lighting, backgrounds, occlusions, and resolutions, thanks to the above-designed simulation in Unity3D.

\footnotetext{Note that this is shown as statistics reported in each paper, but all of them are extendable to generate more data.}

A comparison of RandPerson with other synthetic person re-identification datasets is shown in Table \ref{tab:dataset_compare}. Besides having many more characters and bounding boxes, the most significant feature of RandPerson is that multiple characters move in the scene at the same time, and they are simultaneously captured by multiple non-overlapping cameras, with different viewpoints and backgrounds. In this way, data similar to real surveillance scenarios can be obtained, such as record of transition times in camera networks, and occlusions among people. In contrast, existing synthetic datasets only put a single character in a scene one by one, and capture it using one camera setting after the other. Note that SOMAset in fact has one moving camera configured in 250 different orientations in the same scene. Besides, SyRI captures two videos of two seconds for each person centered in each scene, differing in two different camera orientations. In contrast, videos in the RandPerson are more similar to real surveillance, lasting from tens of minutes to several hours. Also note that, due to labor cost in availability and configuration, the proposed work has much fewer scenes compared to SyRI. However, we are able to explore significantly more variations in the same scene with the camera network, multi-person movements with path navigation, and dynamic lighting.

\section{Experiments}
\subsection{Datasets}

Four widely used real-world person re-identification datasets are used in our experiments, including CUHK03 \cite{li2014deepreid}, Market-1501 \cite{zheng2015scalable}, DukeMTMC-reID \cite{zheng2017unlabeled}, and MSMT17 \cite{wei2018person}. The CUHK03 dataset contains 14,097 images of 1,467 identities. Following the CUHK03-NP protocol \cite{zhong2017re}, it is divided into 7,365 images of 767 identities as the training set, and the remaining 6,732 images of 700 identities as the test set. The detected rather than labeled bounding boxes are used in the experiments. The Market-1501 dataset contains 32,668 images of 1,501 identities, captured from six cameras. The training set contains 12,936 images of 751 identities, and the test set contains the remaining 19,732 images of 750 identities. The DukeMTMC-reID dataset contains 36,411 images of 1,812 identities captured from eight cameras, where 16,522 images of 702 identities are used for training, and the remaining 19,889 images of 1,110 identities are used for testing. The MSMT17 dataset is the largest person re-identification dataset to date. It contains 126,441 images of 4,101 identities, captured from 15 cameras, and a wide range of scenes captured at various time stages. It is divided into 32,621 images of 1,041 identities for training, and the remaining 93,820 images of 3,060 identities for testing. Furthermore, three existing synthetic datasets, SOMAset \cite{barbosa2018looking}, SyRI \cite{bak2018domain}, and PersonX \cite{sun2019dissecting}, are applied for comparison. They are listed in Table \ref{tab:dataset_compare} and described in the related work section. As for the proposed RandPerson dataset, considering its huge number of available images, to reduce redundancy and improve the training efficiency, we randomly sample 132,145 images from the 8,000 identities for training.

This paper mainly focuses on direct cross-dataset evaluation without transfer learning \cite{hu2014cross,yi2014deep,Liao-ECCV2020-QAConv}. We use the Cumulative Matching Characteristic (CMC) \cite{Phillips-HFR2} and mean Average Precision (mAP) \cite{sobh2010innovations} as the performance metrics. All evaluations follow the single-query evaluation protocol \cite{farenzena2010person}. 


\subsection{Implementation Details}\label{sec:impl}

To validate the effectiveness of the RandPerson dataset, we apply a basic deep learning model for cross-dataset person re-identification. All experiments are implemented in PyTorch \cite{pytorch}, using an adapted version \cite{zhong2018camstyle} of the Open-Source Person Re-Identification Library (open-reid) \cite{openreid}. The backbone network is ResNet-50 \cite{he2016deep}. The widely used softmax cross-entropy \cite{bishop2006pattern} is adopted as the loss function. Person images are resized to $256\times128$, then a random horizontal flipping is used for data augmentation. The batch size of training samples is 32. Stochastic Gradient Descent (SGD) \cite{goodfellow2016deep} is applied for optimization, with momentum 0.9, and weight decay $5\times10^{-4}$. The learning rate is set to 0.001 for the backbone network, and 0.01 for newly added layers. When the training data involves real-world images, these rates are decayed by 0.1 after 40 epochs, and the training stops after 60 epochs. Otherwise, these rates are decayed by 0.1 after 10 epochs, and the training stops after 20 epochs.

\subsection{Results of Single-Dataset Training}

First, we perform training on each individual dataset. The cross-dataset evaluation results with CUHK03-NP as the target dataset are listed in Table \ref{tab:CUHK03_result}. Interestingly, the proposed RandPerson dataset outperforms all the real-world datasets, with improvements of 1.8\% - 6.8\% in Rank-1, and 0.2\% - 4.9\% in mAP. To the best of our knowledge, this is the first time a synthetic dataset has outperformed real-world datasets in person re-identification. Compared to existing synthetic datasets, RandPerson also outperforms these by an impressively large margin, with 6.0\% - 13.0\% improvements in terms of Rank-1, and 4.6\% - 10.4\% improvements in mAP.

\begin{table}[htbp] 
 \caption{Results (\%) of models trained on various datasets and tested on CUHK03-NP (detected).}
 \begin{center}
  \begin{tabular}{|l|c|c|}
   \hline
   Training dataset     & Rank-1              & mAP \\
   \hline\hline
   CUHK03       & 21.5          & 19.8  \\
  RandPerson + CUHK03 & \textbf{50.4}         & \textbf{44.4}  \\
    \hline\hline
    DukeMTMC-reID& 6.6          & 5.9\\
    Market-1501   & 7.2          & 6.2\\
    MSMT17       & \textbf{11.6}         & \textbf{10.6}  \\
    \hline
    SOMAset& 0.4          & 0.4\\
    SyRI& 4.1          & 3.5\\
    PersonX& 7.4          & 6.2\\
    RandPerson   & \textbf{13.4}         & \textbf{10.8}  \\
    \hline
    RandPerson + DukeMTMC-reID   & 17.4& 13.8  \\
    RandPerson + Market-1501   & 17.1  & 15.3  \\
  RandPerson + MSMT17 & \textbf{19.1}          & \textbf{17.1} \\
   \hline
  \end{tabular}
 \end{center}
 \label{tab:CUHK03_result}  
\end{table}

Table \ref{tab:Market_result} shows results with Market-1501 as the target dataset. In this popular dataset with a larger scale, RandPerson again outperforms all real-world datasets, with the Rank-1 improved by 5.6\%-22.5\%, and mAP by 5.7\%-15.9\%. Note that DukeMTMC-reID and MSMT17 are both large-scale real-world datasets. Although there is a large domain gap between RandPerson and real-world images, our dataset's  enhanced superior performance indicates that we have successfully enhanced the diversity in identities and scenes. Besides, compared to synthetic datasets, the advantage of RandPerson is even more obvious, with improvements of 11.6\% - 51.1\% in Rank-1, and 8.4\% - 27.5\% in mAP. RandPerson outperforms SyRI by a particularly large margin (26.6\% in Rank-1 and 18.0\% in mAP), indicating that our dataset makes more efficient usage of scenes, while the number of identities is also important. Note that SOMAset is for cloth-independent person re-identification, so training alone with it may not be optimal for general person re-identification in our experiments.

\begin{table}[htbp] 
 \caption{Results (\%) of models trained on various datasets and tested on Market-1501.}
 \begin{center}
  \begin{tabular}{|l|c|c|}
   \hline
   Training dataset     & Rank-1              & mAP \\
   \hline\hline
   Market-1501   & 82.2         & 58.9\\
  RandPerson + Market-1501 & \textbf{87.2}         & \textbf{70.9} \\
   \hline\hline
   CUHK03       & 33.1          & 12.9  \\
   DukeMTMC-reID& 44.9          & 19.0\\
   MSMT17        & \textbf{50.0}         & \textbf{23.1}  \\
    \hline
    SOMAset& 4.5          & 1.3\\
    SyRI& 29.0          & 10.8\\
    PersonX& 44.0          & 20.4\\
   RandPerson    & \textbf{55.6}         & \textbf{28.8}  \\
   \hline
   RandPerson + CUHK03 & 58.0         & 32.5 \\
   RandPerson + DukeMTMC-reID & 60.2      & 32.5 \\
  RandPerson + MSMT17 & \textbf{62.3}         & \textbf{35.8} \\
   
   \hline
  \end{tabular}
 \end{center}
\label{tab:Market_result} 
\end{table}

The cross-dataset evaluation results with DukeMTMC-reID as the target dataset are listed in Table \ref{tab:Duke_result}. As can be observed, compared to synthetic datasets, again, RandPerson improves the performance by a large extent, with a 12.2\% - 43.6\% increase in Rank-1, and 9.0\% - 26.1\% increase in mAP. However, as for the real-world datasets, though RandPerson performs much better than CUHK03-NP and Market-1501, it is inferior to MSMT17. One possible reason is that the domain gap between RandPerson and DukeMTMC-reID is much larger than that between MSMT17 and DukeMTMC-reID, and remember that MSMT17 is the largest and most diverse real-world person re-identification dataset.

\begin{table}[htbp] 

 \caption{Results (\%) of models trained on various datasets and tested on DukeMTMC-reID.}
 
 \begin{center}
  \begin{tabular}{|l|c|c|}
   \hline
   Training dataset     & Rank-1           & mAP \\
   \hline\hline
   DukeMTMC-reID & 74.2  & 53.7\\
RandPerson + DukeMTMC-reID& \textbf{79.4}          & \textbf{60.6} \\
   \hline\hline
   CUHK03       & 24.7          & 10.7  \\
   Market-1501  & 34.6          & 18.1\\
   MSMT17       & \textbf{53.9}         & \textbf{31.3}  \\
   \hline
    SOMAset& 4.0          & 1.0\\
    SyRI& 23.7          & 9.0\\
    PersonX& 35.4          & 18.1\\
   RandPerson   & \textbf{47.6}        & \textbf{27.1}  \\
   
   \hline
  RandPerson + CUHK03& 49.3          & 28.2 \\
 RandPerson + Market-1501& 52.1          & 31.0 \\
 RandPerson + MSMT17& \textbf{61.0}          & \textbf{39.8} \\
  
   \hline
  \end{tabular}
 \end{center}
\label{tab:Duke_result} 
\end{table}

Lastly, from the results shown in Table \ref{tab:MSMT_result}, it can be observed that, when tested on MSMT17, RandPerson outperforms all synthetic and real-world datasets. Nevertheless, all results are quite low, indicating that MSMT17 is more diverse and challenging.

\begin{table}[htbp] 

 \caption{Results (\%) of models trained on various datasets and tested on MSMT17.}
 
 \begin{center}
  \begin{tabular}{|l|c|c|}
   \hline
   Training dataset     & Rank-1            & mAP \\
   \hline\hline
   MSMT17 & 60.8          & 30.7\\
  RandPerson + MSMT17 & \textbf{65.0}         & \textbf{36.8} \\
   \hline\hline
   CUHK03       & 11.9          & 2.9  \\
   Market-1501  & 12.4          & 3.6\\
   DukeMTMC-reID& \textbf{18.3}          & \textbf{5.5}\\
   \hline
    SOMAset& 1.4          & 0.3\\
    PersonX& 11.7          & 3.6\\
    SyRI& 16.4          & 4.4\\
   RandPerson   & \textbf{20.1}       & \textbf{6.3}  \\
   
   \hline
  RandPerson + CUHK03 & 20.1          & 6.7 \\
  RandPerson + Market-1501 & 20.8    & 7.3 \\
  RandPerson + DukeMTMC-reID & \textbf{23.5}   & \textbf{8.3} \\
  
   \hline
  \end{tabular}
 \end{center}
\label{tab:MSMT_result} 
\end{table}

\subsection{Results of Dataset Fusion}

Though the proposed synthetic dataset RandPerson outperforms all existing real-world and synthetic datasets in most cases for cross-dataset person re-identification, it is still not as good as MSMT17 when DukeMTMC-reID is used as the target dataset. This is due to the intrinsic domain gap between synthetic and real-world datasets. One way to reduce the domain gap is to fuse synthetic and real-world datasets. Therefore, we perform another round of experiments after dataset fusion. During training, we simply mix the two kinds of datasets together as if they were from a single dataset, and then perform the same training settings as in Section \ref{sec:impl}.

The results for RandPerson and real-world dataset fusion are also reported in Tables \ref{tab:CUHK03_result}-\ref{tab:MSMT_result} \footnote{RandPerson also outperforms all existing synthesized datasets in these fusion experiments, as shown in the Appendix.}. We make three observations. Firstly, the models trained on real-world datasets are all improved by additionally including RandPerson when training for cross-dataset evaluation. In this case, as reported in Tables \ref{tab:CUHK03_result}-\ref{tab:MSMT_result}, the Rank-1 is improved by 5.2\%-24.9\%, and the mAP is improved by 2.8\%-19.6\%. This indicates that RandPerson, though a synthetic dataset, is complementary to real-world datasets for person re-identification training.

Secondly, even in the same domain, it can be observed that additionally including RandPerson for training considerably improves the performance compared to only using the training data of the target dataset. For example, from the results shown in the first two rows of Table \ref{tab:CUHK03_result}, it can be observed that, compared to training only on CUHK03, the Rank-1 score is significantly improved by 28.9\% and mAP by 24.6\% when additionally including RandPerson.

Lastly, compared to only using RandPerson for training, data fusion also improves the performance by a lot. It can be inferred from the tables that the improvements for Rank-1 range from 0.0\%-13.4\%, and 0.4\%-12.7\% for mAP. This confirms that when real-world datasets are combined with synthetic datasets for training, the domain gap between them can be considerably reduced.

\subsection{Analysis}

\begin{figure}[ht]

\centering
\includegraphics[scale=0.34]{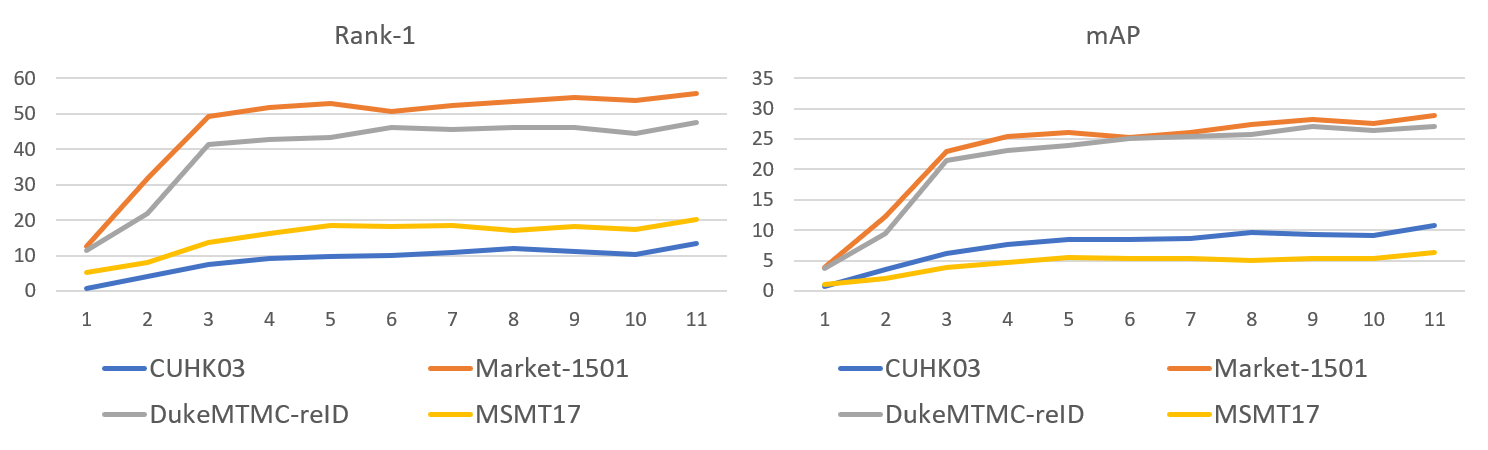}
\caption{Performance of gradually including more scenes.}
\label{fig:scenes_performance}
\end{figure}

To simulate factors that commonly impact re-identification performance, such as variations in viewpoints, poses, lighting, backgrounds and resolutions, we included several relevant and diverse scenes in RandPerson. 
To understand how performance changes when including more scenes, we conduct an experiment in which more scenes are gradually introduced into RandPerson for training. Note that the same 8,000 characters are imported in each scene. Fig. \ref{fig:scenes_performance} shows the effect on the Rank-1 and mAP scores. We can see that most of the scores increase progressively as the number of scenes increases, indicating that different scenes are generally complementary to each other. However, saturation may be reached after a certain number of scenes, and introducing more simply costs additional human labor.

\begin{figure}[ht]

\centering
\includegraphics[scale=0.34]{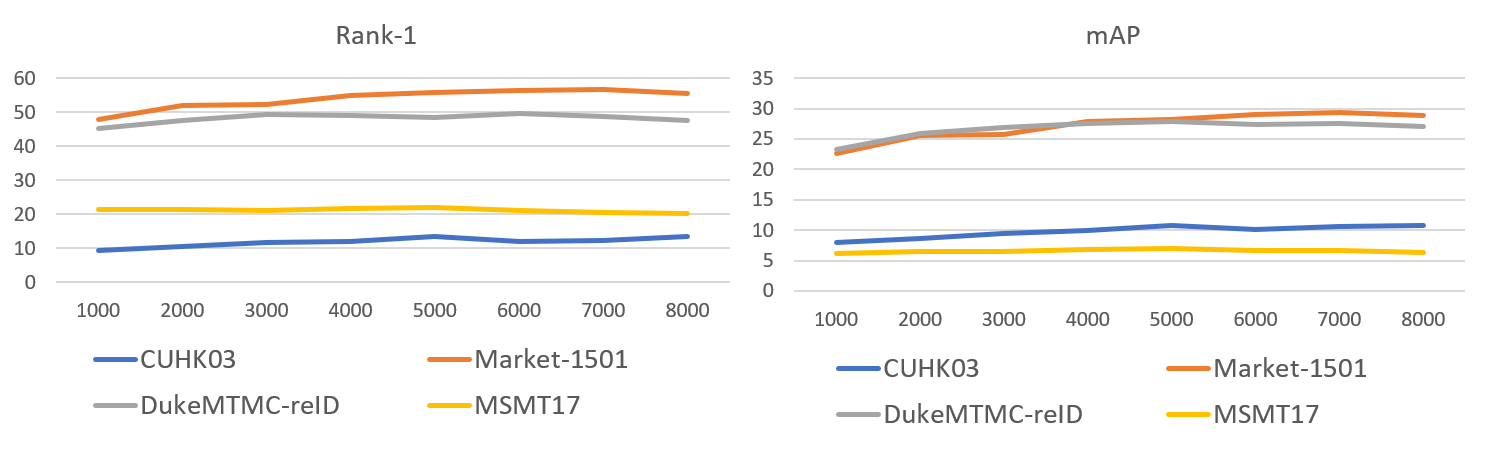}
\caption{Performance with increasing characters.}
\label{fig:ids_performance}
\end{figure}

In addition to the environmental factors that affect data, the number of different identities is also an essential factor influencing person re-identification. Generally, the performance improves as the number of identities increases. Accordingly, we carry out another experiment in which we gradually include more characters in RandPerson for training. Note that no matter how many characters are used, they are all imported to the same 11 scenes. Fig. \ref{fig:ids_performance} shows the effect on the Rank-1 and mAP scores. We can see that the scores gradually increase as the number of characters increases when evaluated on CUHK03-NP, Market-1501 and DukeMTMC-reID. However, when evaluated on MSMT17, the performance of the trained model slowly decreases. This may be because the domain gap between RandPerson and MSMT17 is the largest among all, and hence using more data for training does not help but instead gradually leads to overfitting.

\section{Conclusion}

To improve the generalization of a learned person re-identification model to unknown target domains, this paper contributes a large-scale synthetic dataset called RandPerson. 
This was achieved by creating an automatic code to generate a large number of random 3D characters, and a simulation similar to real surveillance scenarios where multiple persons are moving simultaneously through camera networks. We also used the synthesized person images to train a robust person re-identification model. With this, we proved, for the first time, that a model learned on synthetic data can generalize well to unseen target images, surpassing models trained from many real-world datasets including CUHK03, Market-1501, DukeMTMC-reID, and almost MSMT17. Note that, with the proposed method, more characters and scenes can easily be generated. Besides, the dataset does not require manual annotation, and the privacy issues of real-world data can be avoided. Since the proposed data also includes videos and keypoint recordings, in the future, other tasks can also be explored, such as pose detection and person tracking.

\begin{acks}
This work was partly supported by the NSFC Project \#61672521. The authors would like to thank Anna Hennig who helped proofreading the paper. Some illustration figures in this paper are created with additional public resources as labeled, thanks to the MakeHuman and Unity3D Asset Store where permissions are provided for free usage.
\end{acks}

\bibliographystyle{ACM-Reference-Format}
\bibliography{sample-base}

\title{Appendix}

\makeatletter
\def\maketitle{%
  \begin{center}
    {\Huge \textbf \@title\par}%
    \vspace{5mm}
  \end{center}%
  }
\makeatother

\onecolumn
\maketitle

\setcounter{section}{0}
\setcounter{table}{0}

\section{Full Comparison of Dataset Fusion}

We also fuse SOMAset, SyRI and PersonX with real-world databases, respectively, and the results are reported in Tables \ref{tab:fusion-cuhk}-\ref{tab:fusion-msmt}. From the results, we can see that compared to existing synthetic datasets, RandPerson improves the performance by a large extent, with improvements of 0.2\% - 28.8\% in Rank-1, and 1.1\% - 19.3\% in mAP.

\begin{table}[htbp] 
 \caption{Results (\%) of models trained on various datasets and tested on CUHK03-NP (detected).}\label{tab:fusion-cuhk}
 \begin{center}
  \begin{tabular}{|l|c|c|}
   \hline
   Training dataset     & Rank-1              & mAP \\
   \hline\hline
   SyRI+ CUHK03 & 31.5         & 28.2  \\
  SOMAset + CUHK03 & 35.6         & 31.6  \\
  PersonX + CUHK03 & 43.5         & 40.7  \\
  RandPerson + CUHK03 & \textbf{50.4}         & \textbf{44.4}  \\
    \hline
    PersonX + Market-1501   & 8.2  & 7.7  \\
    SyRI + Market-1501   & 8.9  & 7.8  \\
    SOMAset + Market-1501   & 9.6  & 8.5  \\
    RandPerson + Market-1501   & \textbf{17.1}  & \textbf{15.3}  \\
    \hline
    SOMAset + DukeMTMC-reID   &7.5& 5.9  \\
    PersonX + DukeMTMC-reID   & 8.5& 7.2  \\
    SyRI + DukeMTMC-reID   & 8.6& 7.5  \\
    RandPerson + DukeMTMC-reID   & \textbf{17.4}& \textbf{13.8}  \\
 
  \hline
   SOMAset + MSMT17 & 9.6          & 8.9 \\
  PersonX + MSMT17 & 12.3          & 11.0 \\
     SyRI + MSMT17 & 12.9          & 11.3 \\
  RandPerson + MSMT17 & \textbf{19.1}          & \textbf{17.1} \\
   \hline
  \end{tabular}
 \end{center}
\end{table}

\begin{table}[htbp] 
 \caption{Results (\%) of models trained on various datasets and tested on Market-1501.}\label{tab:fusion-market}
 \begin{center}
  \begin{tabular}{|l|c|c|}
   \hline
   Training dataset     & Rank-1              & mAP \\
   \hline\hline
   SOMAset + CUHK03 & 33.0         & 13.3 \\
   SyRI + CUHK03 & 46.1         & 21.1 \\
   PersonX + CUHK03 & 44.8         & 22.1 \\
  RandPerson + CUHK03 & \textbf{58.0}         & \textbf{32.5} \\
    \hline
   SyRI + Market-1501 & 83.9         & 63.2 \\
   SOMAset + Market-1501 & 82.9         & 63.6 \\
   PersonX + Market-1501 & 82.7         & 65.7 \\
  RandPerson + Market-1501 & \textbf{87.2}         & \textbf{70.9} \\
   \hline
    SOMAset + DukeMTMC-reID & 41.1      & 16.5 \\
    PersonX + DukeMTMC-reID & 45.7      & 21.7 \\
   SyRI + DukeMTMC-reID & 48.9      & 22.1 \\
  RandPerson + DukeMTMC-reID & \textbf{60.2}      & \textbf{32.5} \\
   \hline
   SOMAset + MSMT17 & 41.4         & 18.9 \\
   PersonX + MSMT17 & 43.3         & 21.7 \\
   SyRI + MSMT17 & 49.0         & 23.6 \\
   RandPerson + MSMT17 & \textbf{62.3}         & \textbf{35.8} \\

   \hline
  \end{tabular}
 \end{center}
\end{table}

\begin{table}[htbp] 

 \caption{Results (\%) of models trained on various datasets and tested on DukeMTMC-reID.}\label{tab:fusion-duke}
 
 \begin{center}
  \begin{tabular}{|l|c|c|}
   \hline
   Training dataset     & Rank-1           & mAP \\
   \hline\hline
   SOMAset + CUHK03& 20.5          & 8.9 \\
    PersonX + CUHK03& 32.0          & 16.7 \\
 SyRI + CUHK03& 39.0          & 19.4 \\
 RandPerson + CUHK03& \textbf{49.3}          & \textbf{28.2} \\
   \hline
    SOMAset + Market-1501& 33.8          & 18.1 \\
    PersonX + Market-1501& 36.4          & 19.8 \\
 SyRI + Market-1501& 44.2          & 24.3 \\
 RandPerson + Market-1501& \textbf{52.1}          & \textbf{31.0} \\
   \hline
  PersonX + DukeMTMC-reID& 72.4          & 52.7 \\
SOMAset + DukeMTMC-reID& 73.6          & 53.0 \\
SyRI + DukeMTMC-reID& 74.5          & 54.5 \\
RandPerson + DukeMTMC-reID& \textbf{79.4}          & \textbf{60.6} \\
   \hline
 SOMAset + MSMT17& 47.5         & 26.7 \\
 PersonX + MSMT17& 48.1          & 27.5 \\
 SyRI + MSMT17& 53.9          & 30.6 \\
 RandPerson + MSMT17& \textbf{61.0}          & \textbf{39.8} \\
   \hline
  \end{tabular}
 \end{center}
\end{table}

\begin{table}[htbp] 

 \caption{Results (\%) of models trained on various datasets and tested on MSMT17.}\label{tab:fusion-msmt}
 
 \begin{center}
  \begin{tabular}{|l|c|c|}
   \hline
   Training dataset     & Rank-1            & mAP \\
   \hline\hline
   
    SOMAset + CUHK03 & 9.0          & 2.4 \\
    PersonX + CUHK03 & 11.9          & 3.7 \\
 SyRI + CUHK03 & 19.9          & 5.6 \\
 RandPerson + CUHK03 & \textbf{20.1}          & \textbf{6.7} \\
   \hline
   PersonX + Market-1501 & 10.5    & 3.0 \\
    SOMAset + Market-1501 & 10.7    & 3.2 \\
  SyRI + Market-1501 & 17.4    & 4.2 \\
  RandPerson + Market-1501 & \textbf{20.8}    & \textbf{7.3} \\
   \hline
   SOMAset + DukeMTMC-reID & 15.7   & 4.5 \\
  PersonX + DukeMTMC-reID & 14.8   & 4.6 \\
  SyRI + DukeMTMC-reID & 21.4   & 6.2 \\
  RandPerson + DukeMTMC-reID & \textbf{23.5}   & \textbf{8.3} \\
   \hline
   SOMAset + MSMT17 & 54.3         & 26.4 \\
   PersonX + MSMT17 & 54.0         & 26.8 \\
   SyRI + MSMT17 & 58.6         & 29.4 \\
  RandPerson + MSMT17 & \textbf{65.0}         & \textbf{36.8} \\
   \hline
  \end{tabular}
 \end{center}
\end{table}

\end{document}